\documentclass[11pt]{article}

\usepackage[preprint]{acl}

\usepackage{times}
\usepackage{latexsym}

\usepackage[T1]{fontenc}

\usepackage[utf8]{inputenc}

\usepackage{microtype}

\usepackage{inconsolata}

\usepackage{graphicx}

\usepackage{listings}
\usepackage{xcolor}
\definecolor{darkblue}{rgb}{0, 0, 0.5}
\definecolor{darkgreen}{RGB}{50,100,0}
\definecolor{darkred}{RGB}{200, 0, 0}
\definecolor{gray}{rgb}{0.5,0.5,0.5}
\usepackage{colortbl}
\usepackage{caption}
\usepackage{subcaption}

\usepackage{amsmath}
\usepackage{amssymb}
\usepackage{mathtools}
\usepackage{amsthm}

\usepackage{booktabs} 
\usepackage{comment}

\title{Many Minds from One Model: \\ Bayesian-Inspired Transformers for Population Diversity}

\author{Diji Yang and Yi Zhang\\
  University of California Santa Cruz \\
  \texttt{\{dyang39, yiz\}@ucsc.edu}}

\begin{document}
\maketitle
\begin{abstract}
Despite their scale and success, modern transformers are usually trained as single-minded systems: optimization produces a deterministic set of parameters, representing a single functional hypothesis about the data. Motivated by the analogy to human populations, in which population-level intelligence emerges from diverse individual behaviors, we propose \textbf{Population Bayesian Transformers (B-Trans)}, which enable sampling diverse yet coherent transformer large language model instances (hereafter referred to as a ‘mind’) from a single pre-trained LLM.

B-Trans introduces a Bayesian-inspired posterior proxy by injecting stochasticity directly into normalization layers, avoiding the prohibitive cost of training full Bayesian neural networks. Sampling from this proxy yields a population of \textit{minds} with diverse behaviors while maintaining general competence. During the generation of each response, we sample a single realization from the random distribution and hold it fixed, ensuring temporal consistency and reasoning coherence. Experiments on zero-shot generation and Reinforcement Learning with Verifiable Rewards (RLVR) demonstrate that B-Trans effectively leverages the stochastic model diversity, yielding superior response diversity while achieving better task performance compared to deterministic baselines.
\end{abstract}

\section{Introduction}
The predominant paradigm in Large Language Model (LLM)~\cite{yang2025qwen3,liu2025deepseek} deployment treats model weights as static, deterministic point estimates. While this approach effectively leverages the knowledge acquired during pre-training~\cite{wang2024mmlu}, it inherently limits the model's ability to generate diverse content, i.e., mode collapse~\cite{kirk2023understanding}. Recent studies further demonstrate that well-tuned models suffer from severe structural homogenization~\citep{jiang2025artificial}. Under the convergence pressures of post-training alignment, such as Supervised Fine-Tuning (SFT) and Reinforcement Learning from Human Feedback (RLHF), models tend to collapse (converge) onto a narrow set of reasoning patterns. As a result, the model struggles to explore alternative reasoning paths, locking itself into local optima~\cite{yue2025does}. This rigidity, characterized by strong exploitation but limited exploration, can be viewed as a fundamental bottleneck in complex problem solving and learning for Artificial Intelligence~\cite{sutton1998reinforcement}.

From the perspective of human intelligence, a population of decentralized \textit{minds} can explore more insightful ideas than any single individual, a phenomenon known as the \textit{wisdom of crowds}~\cite{surowiecki2005wisdom}. However, replicating this with a group of models is non-trivial, limited not only by high computational costs but also by the structural homogenization across model instances~\cite{jiang2025artificial}.
Current attempts to reintroduce diversity at inference time operate mostly through injecting randomness into the action distribution, for example, through temperature-based sampling and related entropy-increasing strategies~\cite{wang2022self}. While increasing temperature can prevent verbatim repetition, it often acts as a surface-level perturbation. 
As noted by~\citet{yue2025does}, in reasoning tasks, sampling can lead to incoherent degeneration, yielding statistical variation without meaningful semantic progress, rather than systematically discovering novel solution paths. The core issue is that manipulating the output tokens does not alter the underlying deterministic reasoning engine.

Can we achieve meaningful population diversity without abandoning the efficiency of a single model? In this work, we demonstrate the feasibility by generating many model instances by modifying a single pre-trained model. Our key insight is that normalization layers~\cite{ba2016layer, zhang2019root} modulate representation geometry, and that small shifts, e.g., bias-like offsets, can induce diverse behaviors while preserving semantic meaningfulness.
We therefore replace point-estimated normalization biases with Gaussian-distributed stochastic variables, transforming a single transformer to act as a population of model instances. Each LLM drawn from this distribution corresponds to a distinct \textit{mind}\footnote{In this paper, we use the term \textit{mind} interchangeably with \textit{sampled model instance}. We adopt this terminology to emphasize that each instance acts as a distinct individual from a population.} that shares the same weights and structure, yet processes information through a different normalization parameter (i.e. due to different sampled offsets). This targeted population modeling yields diversity where it is influential, without incurring the cost of maintaining multiple full models.

We present B-Trans as a general method for sampling populations of transformer-based large language models, where each sample represents an individual \textit{mind} (Section~\ref{sec:b-trans}). In this paper, we develop B-Trans through normalization-layer modification (Section~\ref{sec:implementation}), which is deliberately minimal and can be extended to other kinds of modifications in future work. To maintain logical coherence, we maintain temporal consistency by using the same sampled LLM when generating a response sequence. Unlike per-token noise injection (e.g., standard Dropout), which can lead to disjointed logic, our approach ensures that each response is produced by a single, logically consistent model instance sampled from the population, without changing the mind in the middle of a response.

B-Trans is validated through two distinct experimental lenses, each demonstrating different benefits of population diversity. 
In zero-shot generation, we demonstrate that B-Trans effectively breaks structural homogenization, producing semantically meaningful, heterogeneous outputs. 
In Reinforcement Learning with Verifiable Rewards (RLVR)~\cite{guo2025deepseek}, we show that the population-based B-Trans enables deep exploration, traversing sparse reward landscapes more effectively than deterministic model baselines. 
Collectively, these results confirm that B-Trans alters the exploratory dynamics of the model, enabling deep and population-driven reasoning at negligible computational cost.

\section{Related Works}
\label{sec:related_work}

\paragraph{Bayesian Neural Networks and Ensembles.}
Bayesian methods provide a principled framework for modeling weight uncertainty~\citep{blundell2015weight}. 
Prior work has adapted Bayesian principles to neural networks for specific pragmatic objectives. 
For instance, \citet{kristiadi2020being} demonstrated that being ``a bit Bayesian'' (e.g., applying Laplace approximation only to the last layer) can improve calibration and reduce overconfidence (``knowing what you don't know''). 
Similarly, \citet{jing2025kalman} introduced the Kalman Bayesian Transformer to support stable sequential learning and mitigate catastrophic forgetting by treating pre-trained weights as priors. 
While these works primarily use Bayesian tools for calibration, robustness, uncertainty quantification, or model averaging for prediction quality, B-Trans targets a generation-centric goal: we use lightweight parameter-space stochasticity to promote \emph{exploration} and semantic diversity during generation. 
Compared to Deep Ensembles~\citep{lakshminarayanan2017simple}, which scale linearly in cost, or full BNNs, which are computationally intractable, B-Trans induces this population diversity solely through additive modification of normalization computing, offering a negligible alternative specifically optimized for modern LLMs.

\paragraph{Diversity and Mode Collapse in LLMs.}
Standard decoding strategies, such as Temperature~\cite{ficler2017controlling}, Top-$k$~\cite{fan2018hierarchical}, and Nucleus Sampling~\cite{holtzman2019curious}, primarily manipulate the token distribution to include diversity in model output. For example, methods that aggregate responses sampled via these strategies~\cite{wang2022self} can provide slightly different outputs but still remain constrained by intra-mode collapse~\cite{kirk2023understanding}, i.e., a single model generates outputs with low semantic diversity due to the point-estimation nature of the underlying model. 

Another line of research improves the answer diversity in some tasks by ensembling responses from multiple LLMs (population) at a considerable computational cost~\cite{jiang2023llm}. However, recent work points out that modern instruction-tuned models exhibit severe inter-model homogeneity~\cite{jiang2025artificial}, where different LLMs independently converge on similar ideas with minor variations in phrasing. 

B-Trans instead diversifies behavior by sampling model instances via parameter-space modification, turning a single set of weights into a population that can produce diverse generations.

\paragraph{Exploration in Reinforcement Learning.}
Balancing the tradeoff between exploration and exploitation during learning is a long-standing challenge for learning~\cite{sutton1998reinforcement}. Although huge efforts have been made by action-space entropy regularization during training~\cite{haarnoja2018soft,schulman2017proximal}, in practice,~\citet{yue2025does} observe that existing RL methods often force the model into incoherent degeneration (statistical noise) rather than systematically discovering alternative solution paths. 
Meanwhile, as aforementioned, empirical analysis also reveals the structural homogenization of well-aligned models, where the probability mass over-concentrates on a narrow set of reasoning patterns~\cite{jiang2025artificial}.

Alternatively, instead of addressing this intractable optimization problem, we attempt to improve output diversity during the decoding stage. We revisit the concept of Parameter Space Noise~\cite{plappert2018parameter}, originally proposed for continuous control, and adapt it to modern LLMs. B-Trans keeps the pretrained model weight and applies modifications to a lightweight posterior proxy during model inference. The goal is to achieve better exploration via diverse reasoning trajectories.

\section{Bayesian-Inspired Population Model}
\label{sec:b-trans}
This section formalizes a \emph{population} view of a single pre-trained model via Bayesian hypothesis sampling: we draw a sequence-level parameter sample to obtain one coherent model instance, and aggregate multiple individual models for population-level thinking.\footnote{Throughout, ``Bayesian'' is used as a principled motivation for structured uncertainty and sampling, rather than a claim of exact posterior inference for billion-parameter LLMs.}
Section~\ref{sec:implementation} then presents a minimal instantiation that perturbs normalization offsets as a computationally negligible proxy for posterior sampling.

\subsection{Local Geometry and the Laplace Approximation}

Let $\mathcal{D}$ be the training dataset, and $\theta \in \mathbb{R}^d$ be the parameters of a Transformer. We view the pre-trained parameters $\theta_{\text{MAP}}$ as the Maximum A Posteriori (MAP) estimate of the posterior $p(\theta|\mathcal{D})$. While $\theta_{\text{MAP}}$ captures a high-probability configuration, it collapses the rich information contained in the geometry of the loss landscape into a single point. To recover uncertainty, we employ the Laplace Approximation, modeling the posterior proxy around the mode as a Gaussian:

\begin{equation}
    p(\theta|\mathcal{D}) \approx \mathcal{N}(\theta; \theta_{\text{MAP}}, \Sigma).
\end{equation}

Computing the exact covariance $\Sigma$ (the inverse Hessian) is not feasible for billion-parameter models. Furthermore, the true posterior landscape of LLMs is highly non-isotropic \citep{li2018visualizing}, and estimating its anisotropy at inference time without access to training data or gradients is not well-posed.

In principle, constructing the exact posterior geometry presents two pathways: (1) performing full Bayesian training to learn the parameter distribution from vast data associated with different LLMs (text-author pairs), or (2) computing the inverse Hessian matrix.
However, both approaches are computationally prohibitive and data-intensive for modern LLMs. 
We therefore opt for a tractable approximation by empirically defining an isotropic covariance structure $\Sigma = \sigma^2 I$. 
From an information-theoretic perspective, this choice represents the Maximum Entropy distribution given a constraint on the expected Euclidean distance from the $\theta_{\text{MAP}}$ estimate. 
Rather than risking a mis-specified diagonal approximation that might bias the model into degenerate subspaces, the isotropic assumption effectively defines a spherical trust region for exploration. 
Importantly, our goal is not exact Bayesian inference, but efficient sampling from a locally valid posterior proxy, and a practical balancing of theoretical motivation with engineering feasibility.
While this variance is treated as a structural hyperparameter rather than a learned parameter, we show that this simplification is empirically sufficient to induce meaningful population diversity without the intractable costs of the full Bayesian alternative

\subsection{Temporal Consistency: Token-wise vs. Hypothesis Sampling}
\label{sec:temporal_consistency}

A key challenge for Bayesian sampling in autoregressive generation is handling the time dimension. Standard stochastic regularizers, such as Dropout \citep{srivastava2014dropout}, are typically applied independently at each forward pass. For a sequence $y_{1:T}$, this implies integrating over the parameters at every token step:

\begin{equation}
    \label{eq:token_wise}
    p(y_{1:T}|x) \approx \prod_{t=1}^T \int p(y_t|y_{<t}, x, \theta_t) p(\theta_t|\mathcal{D}) \, d\theta_t.
\end{equation}
We argue that Eq.~\eqref{eq:token_wise} is misaligned with coherent multi-step reasoning: it re-samples $\theta_t$ at each token, which can disrupt cross-token logical consistency by changing the underlying computation mid-generation.
Instead, we adopt \emph{hypothesis sampling}. In the Bayesian view, a single sample $\theta \sim p(\theta|\mathcal{D})$ corresponds to one coherent hypothesis. Once sampled, this hypothesis should remain fixed for the entire sequence. 
This also fits the real-world scenario, as each sequence is normally written by one person (which corresponds to one LLM in our sample).
The predictive distribution is:

\begin{equation}
    \label{eq:hypothesis_sampling}
    p(y_{1:T}|x) = \int \left( \prod_{t=1}^T p(y_t|y_{<t}, x, \theta) \right) p(\theta|\mathcal{D}) \, d\theta.
\end{equation}
We approximate Eq.~\eqref{eq:hypothesis_sampling} using Monte Carlo integration with $K$ samples. For each sample $k$, we draw a noise vector $\epsilon^{(k)}$ once and freeze it, obtaining a specific model member $\theta^{(k)} = \theta_{\text{MAP}} + \epsilon^{(k)}$. This ensures that the entire sequence $y_{1:T}$ is generated by a single consistent LLM instance, while enabling population-level aggregation without storing multiple full sets of weights.

\begin{figure*}[ht]
    \centering
        \begin{lstlisting}[
        language=Python,
        basicstyle=\ttfamily\small,
        keywordstyle=\color{blue}\bfseries,
        commentstyle=\color{darkgreen}\itshape, % Comment style in green
        stringstyle=\color{darkred},
        backgroundcolor=\color{white},
        frame=single,
        rulecolor=\color{black},
        numbers=left,
        numberstyle=\tiny\color{gray},
        numbersep=8pt,       % line number size
        xleftmargin=20pt,    % left margin
        framexleftmargin=16pt, % left margin (for line number)
        breaklines=true,
        captionpos=b,
        label={lst:btrans_algo}
    ]
class BayesianBiasWrapper(nn.Module):
    """Augments normalization with a Bayesian latent bias."""
    def __init__(self, base_norm, prior_mu, prior_std):
        self.norm = base_norm
        self.mu, self.sigma = prior_mu, prior_std # z ~ N(mu, sigma)
        self.z = None # cached per sequence

    def forward(self, x):             # x: [B, T, D]
        if self.z is None:
            B, T, D = x.shape
            self.z = sample_normal(self.mu, self.sigma, shape=(B, 1, D))
        return self.norm(x) + self.z  # broadcast along T
        
    def reset_posterior(self):
        """Resamples the model instance (e.g., for a new mind)."""
        self.z = None

    def apply_bayesian_transform(model, prior_mu, prior_std):
        """Converts standard RMSNorm into Bayesian layers."""
        for each normalization layer: # locate layer by model.named_modules
            replace it with BayesianBiasWrapper(module, mu, sigma)
        return model
    \end{lstlisting}
    \vspace{-10pt}
    \caption{\textbf{B-Trans implementation (abstracted).} B-Trans operates as a plug-and-play replacement for normalization layers. For each response, it caches a latent offset per sequence and reuses it across autoregressive steps. This approximates drawing a single coherent model instance for each generation pass, enabling diverse yet temporally consistent outputs from one set of weights.}
    \label{fig:algorithm_code}
\end{figure*}

\section{Implementation}
\label{sec:implementation}
Our approach is a surgical architectural modification during the model inference phase. 
Aligning with the pragmatist view that partial Bayesian structure in carefully chosen components can already induce useful responses~\citep{kristiadi2020being}, we implement B-Trans by wrapping RMSNorm layers with a sampled offset. 
B-Trans operates as a plug-and-play replacement for normalization layers, ensuring that the modified LLM is still compatible with existing inference engines such as vLLM~\cite{kwon2023efficient}. Furthermore, B-Trans remains differentiable, which makes backpropagation possible for all kinds of learning in future works. 

\subsection{Bayesian Bias Norm Wrapper}
Recent LLMs typically employ RMSNorm, which is often implemented without an explicit additive bias. 
In B-Trans, we treat the post-normalization additive shift as a bias-like offset and inject a sequence-level latent variable $\mathbf{z}$ after normalization. 

Our decision to target bias-like offsets is motivated by prior findings~\cite{elhage2021mathematical,dong2021attention} that biases encode ``default'' priors and are crucial for rank restoration, allowing us to shift reasoning dispositions without corrupting the semantic correlations stored in weights (see more discussion in Section~\ref{subsec:why_biases}).
Concretely, for a hidden state $\mathbf{x}$, we implement:

\begin{equation}
\mathbf{y}
=
\mathrm{Norm}(\mathbf{x};\mathbf{w},\mathbf{b})
+\mathbf{z},
\quad
\mathbf{z}\sim\mathcal{N}(\boldsymbol{\mu},\,\sigma^{2})
\end{equation}

Here, $\mu$ and $\sigma$ are hyperparameters controlling the center and spread of the population diversity, and $\mathbf{b}$ can be 0 in RMSNorm (which makes $z$ a post-norm additive offset).
For normalization variants that include a learned scale (and optionally a bias), these parameters remain part of the deterministic backbone, while $\mathbf{z}$ provides a stochastic post-normalization offset for each sampled model instance. 

\subsection{Biases as Dispositional Priors}
\label{subsec:why_biases}
B-Trans introduces stochasticity through additive offsets rather than through full weight modification. 
This choice is grounded in the functional role of additive shifts in Transformer mechanics. 
First, additive components act as input-independent defaults: while weight matrices encode correlations between features, additive shifts set baseline writes to the residual stream and influence activation thresholds before input-specific computation~\cite{elhage2021mathematical}. Perturbing these terms, therefore, alters a model’s inductive dispositions without corrupting the semantic relationships stored in its weights.
Second, additive vectors are a practical handle for semantic steering. Works on activation engineering show that adding vectors in the residual stream can steer high-level behaviors without broadly degrading capabilities~\cite{turner2023steering, zou2023representation}. 
Third, bias-like shifts can be critical for model behavior.~\citet{sun2024massive} suggests that when explicit biases are absent, models may simulate similar effects through outlier activations. 
For these reasons, B-Trans targets additive offsets as a minimal and controlled locus for inducing population diversity. While extending stochasticity to additional parameters—such as normalization weights 
$w$ (Equation 4)—may further enrich diversity, we do not explore such extensions in this work. Perturbing full weight matrices of LLMs, in contrast, risks disrupting learned representations and introducing excessive noise, potentially degrading model behavior. We therefore restrict our study to additive offsets and leave broader normalization- or weight-level stochasticity to future work.

\subsection{Temporal Consistency via Noise Caching}
As mentioned in Section~\ref{sec:temporal_consistency}, for each sequence generation, we sample one LLM (analogous to a distinct individual from the population) to solve the problem.
Thus, we implement a \textit{sequence-level} noise caching mechanism (see Figure~\ref{fig:algorithm_code}). As detailed in the \texttt{BayesianBiasWrapper}, the process ensures that the stochastic perturbation defines a stable reasoning flow for the entire response:
(1) Initialization: Upon receiving the prompt, we sample the latent variable $\mathbf{z}$ once and cache it, effectively drawing a single sample $\theta^{(k)}$ from the approximate posterior.
(2) Generation: For all subsequent autoregressive steps within the same sequence, the model applies this frozen $\mathbf{z}$ to the normalization layers.

This design ensures that the stochasticity is strictly inter-sequence rather than intra-sequence. Crucially, unlike full weight-masking methods (e.g., MC Dropout), our implementation is computationally negligible, requiring only an element-wise addition of a cached tensor without the overhead of generating large binary masks. 
Specifically, implementing theoretically correct (frozen) MC Dropout on a 7B model would require caching around 14GB of masks per user instance, assuming pre-expanding the binary masks to FP16 to align with weight precision to avoid prohibitive on-the-fly decompression overheads.
In contrast, B-Trans only caches the bias noise vectors (<1MB), reducing the memory footprint by orders of magnitude while maintaining temporal consistency.

\section{Experiments}
\label{sec:experiments}
To validate that B-Trans can produce diverse but meaningful \textit{minds} from one model, we conduct experiments under two settings:
(1) B-Trans for zero-shot generation, focusing on improving semantic diversity and also pass@k for accuracy, and (2) B-Trans for output sampling, focusing on providing a useful learning signal for Reinforcement Learning with Verifiable Rewards (RLVR). 
Across both settings, our goal is to provide a minimal proof-of-concept for population-based exploration under controlled, standard protocols, rather than an exhaustive benchmark. 

\subsection{Zero-Shot Diversity and Creativity}
\label{subsec:exp_zeroshot}

The premise of B-Trans is that each sampled response corresponds to a different model instance. 
We evaluate whether B-Trans yields meaningful heterogeneity compared to the deterministic baseline with high-temperature decoding.

\begin{figure}[t]
    \centering
    \includegraphics[width=0.98\linewidth]{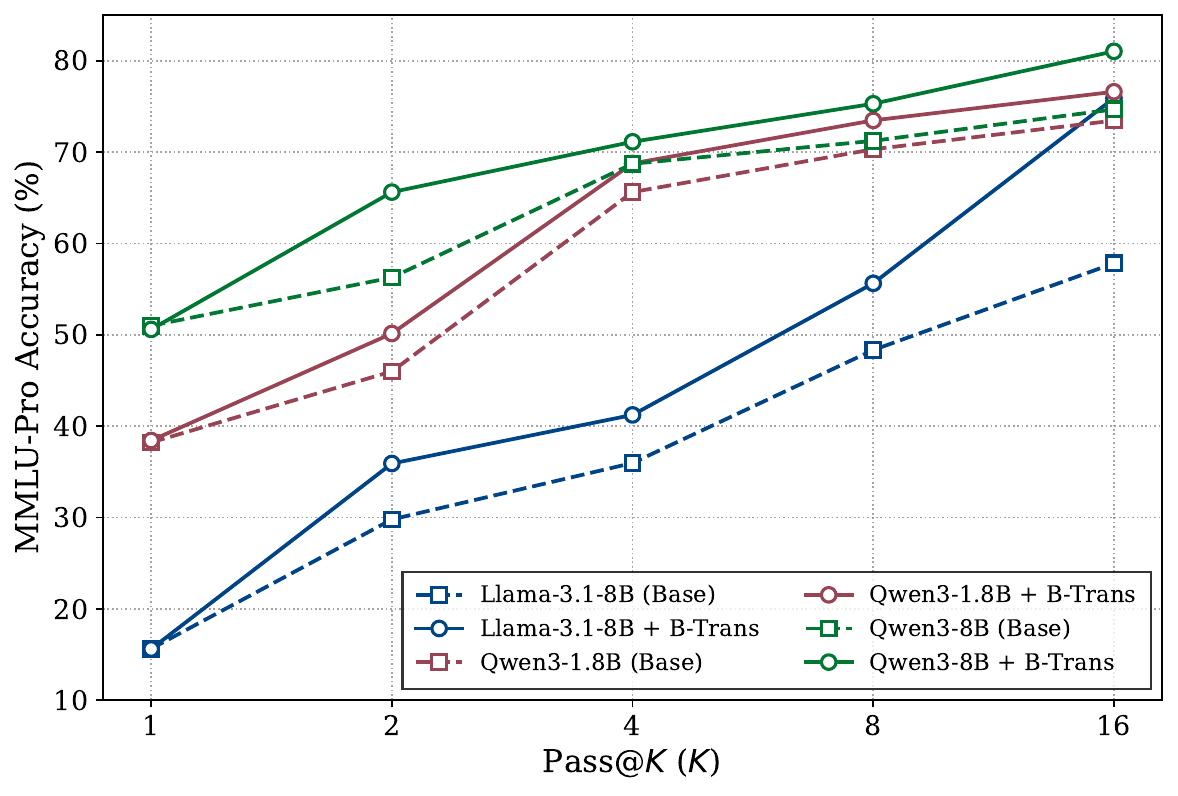}
    \caption{Pass@$K$ performance on MMLU-Pro across Llama-3.1 and Qwen3 families. Solid lines denote B-Trans, while dashed lines represent deterministic baselines with high-temperature sampling. While performance at $K=1$ is comparable, B-Trans exhibits superior scaling at higher $K$ values. This widening gap indicates that the induced parameter-space diversity yields valid, functional reasoning paths rather than random noise.}
    \label{fig:zeroshot-mmlu-pro}
\end{figure}

\begin{figure}[ht]
    \centering
    \includegraphics[width=0.98\linewidth]{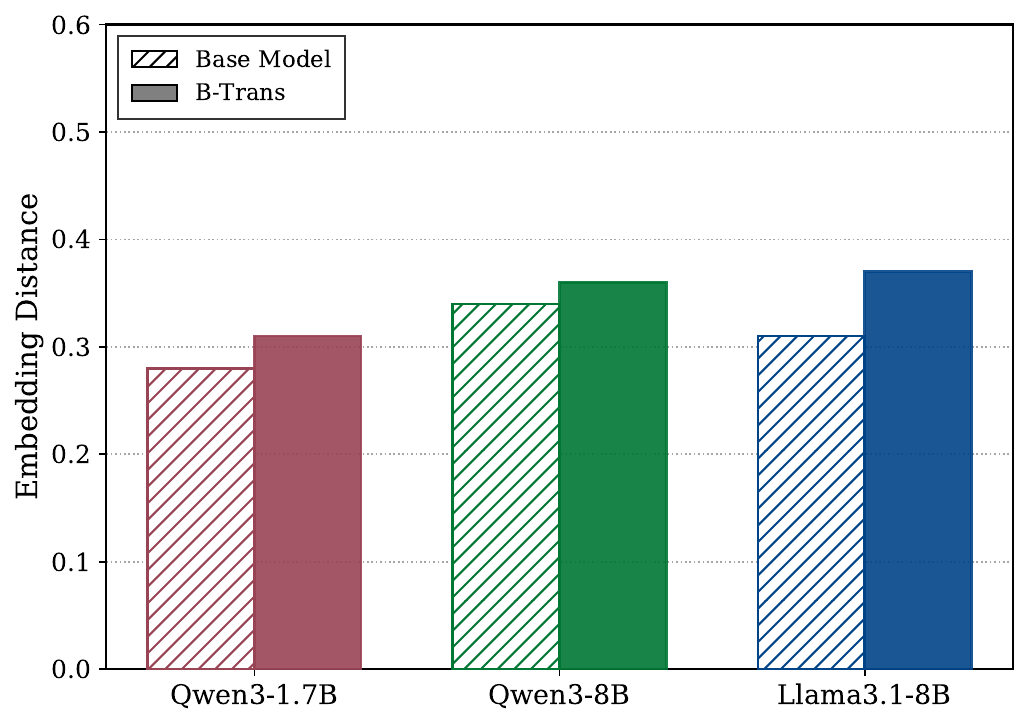}
    \caption{Semantic diversity on INFINITY-CHAT. B-Trans (solid) consistently yields higher embedding distances than baselines (hatched), confirming better output heterogeneity on creativity writing.}
    \label{fig:zeroshot-infinity-chat}
\end{figure}

\begin{figure*}[ht]
    \centering
    \includegraphics[width=0.95\textwidth]{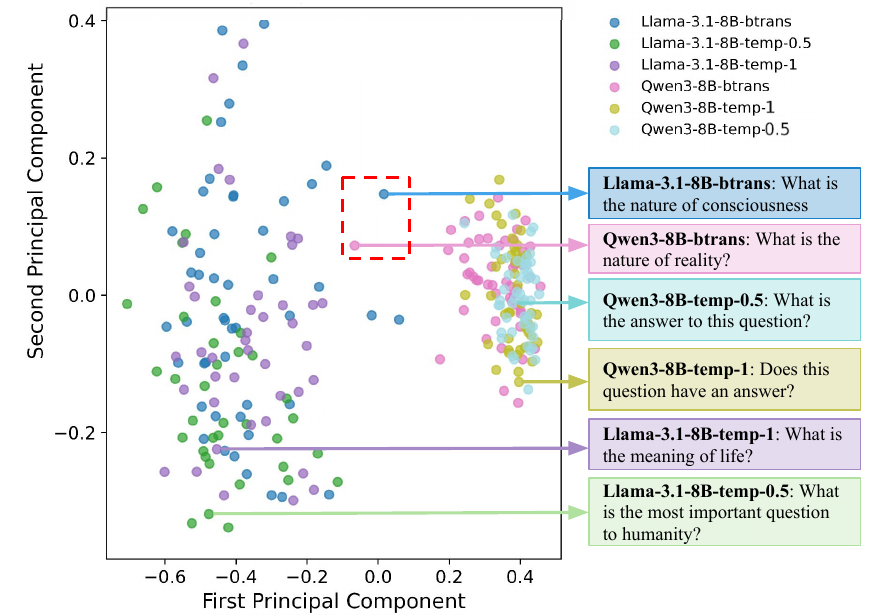}
    \caption{PCA visualization of response embeddings from Qwen3 and Llama-3.1 models. The query is ``\textbf{Output a hard question to humanity (super concise and short), independent of theme.}'' Standard sampling confines models to small semantic regions. B-Trans significantly expands the territories, allowing the model to escape its default mode and even explore semantic regions typically associated with the other models (highlighted in {\color{red}{red}} box).}
    \label{fig:hivemind}
\end{figure*}

\paragraph{Setting.}
We study intrinsic population behavior on two benchmarks with different measurement goals.
On MMLU-Pro~\cite{wang2024mmlu} (a challenging scientific reasoning dataset that contains 12K complex questions across various disciplines), we use Pass@K to test whether additional sampled individuals provide valid reasoning paths: for each question, we sample $K$ independent LLM instances and count the answer as correct if any sample is correct.
Thus, improved scaling with larger $K$ indicates that the induced diversity is not mere repetition but yields a correct solution.
On INFINITY-CHAT~\cite{jiang2025artificial} (100 open-ended questions that admit a wide range of plausible answers with no ground-truth correctness), we measure semantic diversity using average pairwise embedding cosine distance~\cite{reimers2019sentence,song2020mpnet} and visualize the embedding geometry with PCA following Artificial Hivemind analysis~\citep{jiang2025artificial}. The goal is to quantify and visualize semantic heterogeneity.
For both tasks, we contrast B-Trans against two distinct base models, Qwen3 1.7B, 8B~\cite{yang2025qwen3}, and Llama-3.1~\cite{grattafiori2024llama}. All models share the same decoding strategies. 

\paragraph{Results.}
On MMLU-Pro (Figure~\ref{fig:zeroshot-mmlu-pro}), while B-Trans matches the baseline at $K=1$, it demonstrates significantly stronger scaling behavior as the sample budget increases. This trend confirms that the diversity introduced by B-Trans is semantically meaningful: rather than generating redundant repetition (which would yield diminishing returns), the B-Trans population successfully explores valid reasoning trajectories that are not found by the deterministic baselines. 
Complementing this, the embedding distance analysis on INFINITY-CHAT (Figure~\ref{fig:zeroshot-infinity-chat}) further quantifies this behavior, confirming better output heterogeneity.

\paragraph{Qualitative Analysis} 
Figure~\ref{fig:hivemind} visualizes the semantic landscape, where spatial proximity indicates the semantic similarity of response embeddings (dots). Note that the horizontal distance reflects the most fundamental conceptual differences between each response.

Under standard sampling (temp-0.5), Qwen3 and Llama-3.1 occupy disjoint, narrow regions in the semantic space, reflecting their distinct pre-trained biases. 
While high temperature (temp-1) slightly expands these clusters, it fails to alter their fundamental topology. 
In contrast, B-Trans generates a significantly broader distribution, forming distinct clusters that explore diverse reasoning paths. 

Notably, B-Trans can occasionally reach semantic regions that are atypical for the original base model. As examples highlighted in the red box, Qwen3-8B-btrans generates one response that is very close to Llama-3.1's style. This validates our goal of mining different minds from one model.

\subsection{Exploration in Reinforcement Learning with Verifiable Rewards (RLVR)}
\label{subsec:exp_rlvr}
In Reinforcement Learning, the critical bottleneck is often the exploration-exploitation trade-off. Standard LLM policies rely on action-space noise (sampling tokens) to explore. In this section, we show that parameter-space noise (sampling models) provides a more consistent and deeper form of exploration under the setting of RLVR.

\paragraph{Setting.} 
We follow the default configuration of the SimpleRL-Zoo framework as-is, which is a popular RLVR recipe to improve models' reasoning abilities using Group Relative Policy Optimization (GRPO)~\cite{guo2025deepseek}. Our studies include different sizes of Qwen3 models (1.7B and 8B), math-specialized Qwen2.5-Math-7B, and Llama3.1-8B. 
For each model, we compare two approaches to sample a group of responses (16 outputs) for each input for GRPO reward computing : (1) using a deterministic base model, and (2) using B-Trans. Note that we keep the policy model the same for both settings and focus solely on the difference in sampling strategies.
To save computing, we only update Low-Rank Adaptation (LoRA)~\cite{hu2022lora} adapters for the policy model instead of full-parameter fine-tuning (less than 1\% trainable parameters), and we train on 400 examples sampled from the same distribution of the SimpleRL-Zoo dataset via difficulty-stratified sampling.
We test the finetuned model on three standard mathematical reasoning datasets, i.e., GSM8K~\cite{lewkowycz2022solving}, MATH-500~\cite{lightman2023lets}, and Minerva Math~\cite{lewkowycz2022solving}, where the ground truth is verifiable.

\paragraph{Results.} 
As shown in Table~\ref{table:rlvr_results}, B-Trans achieves consistent gains across all four benchmarks.
GRPO relies on informative relative comparisons within the sampled group. When the rollout group is homogeneous, relative advantages become less informative, and updates can stagnate. 
By producing group members from distinct parameter-space instances, B-Trans increases the chance that at least one candidate reaches a higher-reward region, strengthening the learning signal. 

\begin{table}[h]
\centering
\resizebox{\columnwidth}{!}{
\begin{tabular}{lcccc}
\toprule
\textbf{Model} &
\textbf{GSM8K} &
\textbf{MATH-500} &
\textbf{Minerva Math} &
\textbf{Average} \\
\midrule

Qwen3-1.7B & 86.3 & 47.0 & 21.3 & 51.5 \\ 
\rowcolor[rgb]{ .867, .922, .969}~~~~$\hookrightarrow$~B-Trans & 86.8 & 54.4 & 26.1 & 55.7 \\ 
Qwen3-8B & 90.0 & 60.6 & 40.4 & 63.7 \\ 
\rowcolor[rgb]{ .867, .922, .969}~~~~$\hookrightarrow$~B-Trans & 90.7 & 61.4 & 42.6 & 64.9 \\ 
Qwen2.5-Math-7B & 83.6 & 71.2 & 30.5 & 61.8 \\ 
\rowcolor[rgb]{ .867, .922, .969}~~~~$\hookrightarrow$~B-Trans & 88.1 & 76.0 & 32.7 & 65.6 \\ 
Llama3.1-8B & 73.8 & 20.8 & 7.4 & 34.0 \\ 
\rowcolor[rgb]{ .867, .922, .969}~~~~$\hookrightarrow$~B-Trans & 78.6 & 23.7 & 9.9 & 37.4 \\ 

\bottomrule
\end{tabular}
}
\caption{Accuracy (\%) on RLVR tasks using GRPO. B-Trans consistently outperforms the baseline.}
\label{table:rlvr_results}
\end{table}

\section{Ablation Study}
\label{sec:ablation}

To validate the necessity of hypothesis sampling (Section \ref{sec:temporal_consistency}), we compare B-Trans against a \textit{Token-wise Noise} baseline on MATH-500 using Qwen3-1.7B. While B-Trans samples a single latent $z$ per sequence, the baseline resamples $z_t \sim \mathcal{N}(0, \sigma^2I)$ at every forward pass. We fix an identical noise magnitude ($\sigma=0.02$) for both settings to isolate the impact of temporal inconsistency.
 
We use the \textit{Step-wise Consistency Score} (SCS) to quantify the logical stability of the reasoning chain. We segment each Chain-of-Thought $Y$ into discrete steps $S = \{s_1, \dots, s_m\}$ based on newline delimiters. The SCS is defined as the average cosine similarity between embeddings of consecutive steps (Encoded via \texttt{all-MiniLM-L6-v2}):
\begin{equation}
    \text{SCS}(Y) = \frac{1}{m-1} \sum_{i=1}^{m-1} \cos(\text{Enc}(s_i), \text{Enc}(s_{i+1}))
\end{equation}

\paragraph{Results and Analysis} 
The Token-wise baseline suffers catastrophic degradation. Despite identical noise levels, its SCS drops to \textbf{0.42} (vs. \textbf{0.58} for B-Trans), correlating with a decrease in accuracy (40.2\% vs. 46.4\%). This empirical failure corroborates the theoretical insight established by \citet{gal2016dropout} at RNNs/LSTMs age: valid variational inference requires the stochastic mask to remain invariant across time steps to represent a single coherent hypothesis. B-Trans satisfies this Temporal Consistency requisite through lightweight bias perturbation, effectively capturing the benefits of weight uncertainty without the memory overhead of full ensemble masks.

\section{Future Work and Discussion}
\label{sec:future-work}

Our current implementation models the prior distribution as a simple isotropic Gaussian with a scalar hyperparameter $\sigma$. While this serves as an effective proof-of-concept, the assumption of uniform uncertainty across all layers and dimensions is a strong simplification. Furthermore, as mentioned in Section~\ref{subsec:why_biases}, we could also introduce a distribution over parameters other than Bias-like offset, and the weights in normalization could be a natural target.

Another progression for this line of research is to develop a learnable or hierarchical prior. In this initial exploration, we adhered to a fixed $\sigma$ to mitigate the risks associated with post-training optimization. 
Learning variance parameters on limited downstream datasets can lead to overfitting, where the induced population loses its general competence. 
Ideally, the population model can be learned from a large corpus reflecting real population behavior, similar to using real human interaction data for Bayesian recommender system training~\cite{zhang2007efficient}. 
We anticipate that future efforts, particularly those with sufficient computational resources to train on large-scale general corpora, could successfully learn these uncertainty parameters. Such a data-driven prior would likely act as a more robust population, capturing layer-specific sensitivities that a scalar $\sigma$ cannot.

Furthermore, the scalar $\sigma$ presents an opportunity for adaptive inference. The hyperparameter could be meta-learned or dynamically adjusted based on input complexity, e.g., allowing the model to be confident (low variance) on rote tasks and imaginative (high variance) on open-ended problems. Ultimately, we envision a paradigm shift where LLMs are no longer delivered as static artifacts, but as probabilistic distributions capable of adapting their internal state to the nature of the query.

\section{Conclusion}
\label{sec:conclusion}

We revisit the prevailing view of Large Language Models as static point estimates and argue that lightweight parameter-space modification is a practical catalyst for exploration and diversity. Benefited from the \textit{wisdom of crowds}, our experiments demonstrate that B-Trans, even with a simple instantiation, provides meaningful and diverse responses at negligible cost. In an era where foundation models have achieved strong deterministic prediction capabilities (exploitation), we envision that B-Trans contributes to the next generation of models by breaking through structural homogenization and facilitating efficient learning via deeper exploration.

\clearpage

\section*{Limitations}
While B-Trans demonstrates a promising direction for inducing implicit ensembles, our current study is intentionally scoped as a minimal, representative proof-of-concept centered on architectural simplicity and controlled comparisons.
We do not perform extensive hyperparameter sweeps, broad architecture coverage, or large-scale benchmarking beyond validating the key design principles.
Our intervention is restricted to the biased terms of normalization layers, which is a simplified proxy rather than a full Bayesian neural network.
While this surgical approach is computationally negligible, it is a simplified approximation of a full Bayesian Neural Network. The weights in the attention and feed-forward layers remain deterministic, potentially limiting the expressivity of the induced posterior.

\section*{Ethics Considerations and Broader Impact}
This work introduces a parameter-space sampling method to enhance LLM diversity during inference. Regarding potential risks, we acknowledge that techniques improving generative diversity could theoretically be misused to generate varied disinformation or spam. However, our primary objective is to enable deeper exploration in reasoning tasks. Environmentally, B-Trans offers a positive impact by approximating population-level intelligence from a single model, thereby avoiding the significant carbon footprint associated with training and maintaining full model ensembles.

\bibliography{custom}

\end{document}